\pdfoutput=1

\documentclass[11pt]{article}

\usepackage{acl}  

\usepackage{times}
\usepackage{latexsym}

\usepackage[T1]{fontenc}

\usepackage[utf8]{inputenc}

\usepackage{microtype}

\usepackage{inconsolata}

\usepackage{multirow}
\usepackage{array}
\usepackage{graphbox}
\usepackage{color}
\usepackage{graphicx}
\usepackage{comment}
\usepackage{amsmath,amssymb} 
\usepackage{soul}
\usepackage{float}
\usepackage{pifont}
\usepackage{wrapfig}
\usepackage{makecell}

\definecolor{blockyellow}{RGB}{252, 209, 42}
\definecolor{blockviolet}{RGB}{112, 48, 160}
\definecolor{blockred}{RGB}{192, 0, 0}
\definecolor{blockorange}{RGB}{249, 166, 2}

\newcolumntype{?}[1]{!{\vrule width #1}}

\newcommand*{\ShowNotes}{}
\ifdefined\ShowNotes
  \newcommand{\colornote}[3]{{\color{#1}\bf{#2: #3}\normalfont}}
\else
  \newcommand{\colornote}[3]{}
\fi

\newcommand{\ctext}[3][RGB]{%
  \begingroup
  \definecolor{hlcolor}{#1}{#2}\sethlcolor{hlcolor}%
  \hl{#3}%
  \endgroup
}

\title{MedCycle: Unpaired Medical Report Generation via Cycle-Consistency}

\author{Elad Hirsch \hspace{0.2in} Gefen Dawidowicz \hspace{0.2in} Ayellet Tal \\
Technion – Israel Institute of Technology \hspace{0.1in}\\
}

\begin{document}
\maketitle
\begin{abstract}
Generating medical reports for X-ray images presents a significant challenge, particularly in unpaired scenarios where access to paired image-report data for training is unavailable.
Previous works have typically learned a joint embedding space for images and reports, necessitating a specific labeling schema for both. 
We introduce an innovative approach that eliminates the need for consistent labeling schemas,
thereby enhancing data accessibility and enabling the use of incompatible datasets.
This approach is based on cycle-consistent mapping functions that transform image embeddings into report embeddings, coupled with report auto-encoding for medical report generation.
Our model and objectives consider intricate local details and the overarching semantic context within images and reports. 
This approach facilitates the learning of effective mapping functions, resulting in the generation of coherent reports.
It outperforms state-of-the-art results in unpaired chest X-ray report generation, demonstrating improvements in both language and clinical metrics.
\end{abstract}

\begin{figure}[t]
  \begin{center}
    \includegraphics[width=0.99\linewidth]{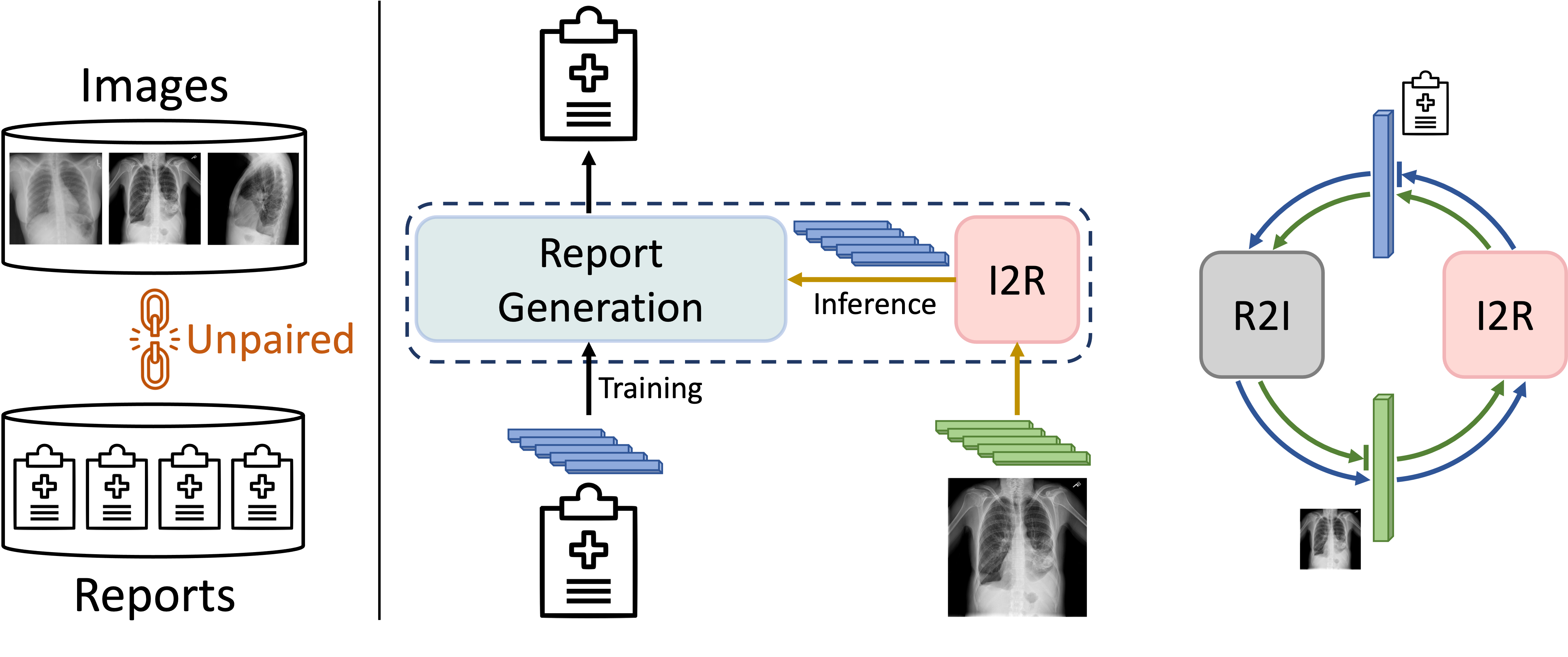}
  \end{center}
{ \small 
\hspace{0.13in}(a) Data \hspace{0.2in} (b) Generation via Cycle-Consistent Mapping
}
\caption{{\bf Unpaired medical report generation.} 
(a)~Two unpaired datasets are available: chest X-ray images and chest X-ray reports.
(b) Our model learns cycle-consistent mappings between image and report embedding spaces ($I2R$ \& $R2I$), facilitated by cross-modality alignment through the use of pseudo-reports, as well as report auto-encoding. 
Report generation is executed by decoding transformed image representations into reports during inference.
}
\label{fig:teaser}
\end{figure}

\section{Introduction}
\label{sec:introduction}

Automating the generation of medical reports has the potential to improve the efficiency of patient information analysis and documentation, leading to better care and cost savings.
Consequently, many research efforts have been directed towards this aim~\cite{chen-acl-2021-r2gencmn,chen-etal-2020-generating,jing2017automatic,li2019knowledge,wang2022cross}.
These works rely on labeled image-report paired datasets~\cite{demner2016preparing,johnson2019mimic}, which are relatively small and less accessible in comparison to datasets for natural images~\cite{lin2014microsoft,sharma2018conceptual,thomee2016yfcc100m}.
Privacy concerns, restricted access to high-quality data, and the complex nature of medical data analysis and labeling, demanding specialized expertise, contribute to this problem and limit the availability of such paired data.
Even when paired datasets exist, they may not be fully accessible to the public, leading to partially-available datasets, such as the one presented in~\cite{irvin2019chexpert}, which consists solely of images.
These limitations in obtaining paired data pose a significant challenge in this domain.

Utilizing unpaired data, i.e., images and reports originating from different sources, may help alleviating some of these limitations.
Specifically, it may resolve privacy or regulatory concerns and increase the available amount of training data.
Nonetheless, when attempting to learn report generation from images, the absence of image-report pairs introduces a significant challenge.
The only previous work that addressed this task involved constructing a knowledge graph of the domain and utilizing a classification module for pathologies~\cite{liu2021auto}.
This process requires expertise in constructing the knowledge graph and, most importantly, ensuring that the images and reports are labeled according to a consistent and shared schema.
These labels typically pertain to various thoracic pathologies.
Our motivation is to eliminate these constraints.

We propose to address the challenge by considering four perspectives, each contributing to a cohesive solution: embedding spaces, mapping between these spaces, initial cross-modality alignment, and report generation (Figure~\ref{fig:teaser}).
Specifically, we construct two separate embedding spaces---one for the visual modality (images) and the other for the textual modality (reports)---utilizing joint global and local representations. 
Subsequently, mapping functions learn the transformation from an image representation to a report representation and vice versa.
As mapping functions should preserve the semantic meaning of the data, and image-report pairs are unavailable, training a mapping function to transform image representation to its corresponding report representation becomes challenging. 
To overcome this issue, we train the mapping between the two embedding spaces to preserve cycle consistency.
Additionally, to establish initial cross-modality relationships, we introduce a novel concept of pseudo-reports, leveraging available domain information accompanying the image dataset (e.g., pathologies). 
We encourage the representation of an input image, after mapping to the report space, to closely align with the representation of its corresponding pseudo-report.
Lastly, a decoder is exclusively trained with reports, utilizing auto-encoding, to generate medical reports.

Overall, at inference time, given an image, we use both the learned mapping to the report domain and the knowledge learned through auto-encoding to generate a report that suits the image.

Our model adeptly handles a fundamental requirement of report generation---the need for details. 
Recall that the available data is solely global, indicating the presence of pathologies and encompassing the entire image or report.
Alignment at this global level often results in overly generalized representations, which may be suitable for classification but fall short in capturing fine details in individual examples---details crucial for effective report generation.
Conversely, local representations---those depicting image patches or individual report words—capture numerous details, providing essential information for the report generation process. However, they lack alignment.

The effectiveness of our method is evident from improvements in both language and clinical metrics.
For instance, when compared to previous SoTA methods, our approach demonstrates a $9\%$ enhancement in the BLEU-1 score (language efficacy) and a $3\%$ increase in F1 (clinical efficacy) on the dataset from~\cite{johnson2019mimic}, all while eliminating the need for specific training dataset requirements. 
Furthermore, we illustrate how the absence of these requirements allows the utilization of other training datasets, resulting in further performance improvements.

Hence, our paper makes the following contributions:
\begin{enumerate}
    \item 
    We introduce a novel approach to generate medical reports from images in an unpaired setting. This approach is based on learning cycle-consistent mapping functions between domains, establishing cross-modality relations through a novel concept of pseudo-reports, and utilizing an auto-encoding model to generate reports from images.
    \item
    Our method eliminates the need for image and report datasets to be labeled with a consistent schema, thereby increasing data accessibility. This enables the utilization of datasets that were previously incompatible due to differences in pathology labels or languages.
    \item
    Our method outperforms the SoTA results in unpaired chest X-ray report generation.
\end{enumerate}

 \begin{figure*}[t]
\centering
\includegraphics[width=1.0\linewidth]{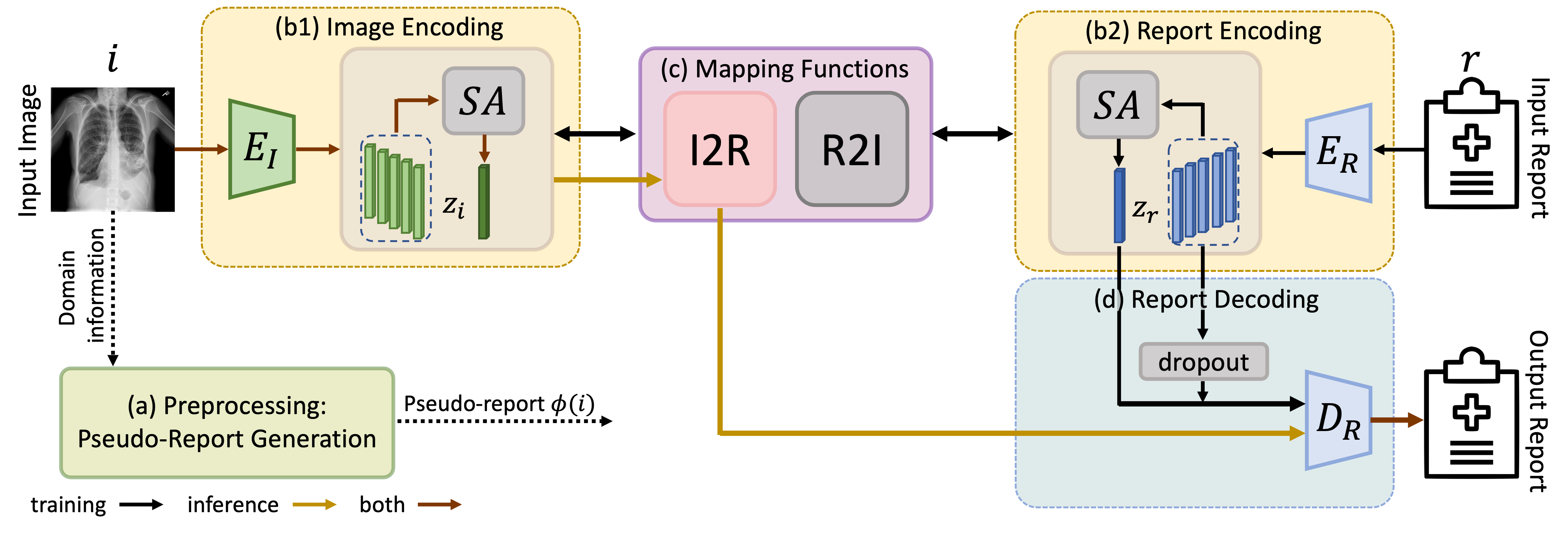}
\caption{{\bf Method.}
For each image $i$ from a dataset $\mathcal{D}_I$, a preprocessing step generates a corresponding pseudo-report denoted as $\phi(i)$~(a), which conveys essential image information in textual form.
An image  encoder encodes each image $i$ into $z_i$~(b1). 
Simultaneously, a report encoder encodes reports from a report dataset $\mathcal{D}_R$~(b2), as well as pseudo-reports. 
These encoded representations comprise both local and aggregated global features, by employing self-attention $SA$.
Two mapping functions are trained
to transform image representations into report representations ($I2R$), and vice versa ($R2I$)~(c) .
Subsequently, a decoder (d) utilizes the encoded reports (excluding pseudo-reports) to output a report, aiming to reconstruct the initial report.
For improved generalization, dropout masks a portion of the local representations.
During inference, an input image is encoded~(b1), followed by mapping to the report space~(c). 
The transformed representation is then decoded to generate a report~(d).
}
\label{fig:method}
\end{figure*}

\section{Related Work}
\label{sec:related}

\noindent
{\bf Paired medical report generation.}
Methods that rely on paired data have access to both images and their corresponding reports.
These models typically employ an encoder-decoder architecture, where the encoder extracts visual features, commonly using a CNN, and the decoder generates text. 
Some models utilize a hierarchical decoder comprising topic and word decoders~\cite{jing2017automatic,liu2019clinically,zhang2020radiology}, while others employ Transformers~\cite{chen-acl-2021-r2gencmn,chen-etal-2020-generating,huang2023kiut,li2023unify,hou-etal-2023-organ}.
Knowledge graphs~\cite{li2019knowledge,liu2021exploring,zhang2020radiology} and memory blocks~\cite{chen-acl-2021-r2gencmn,chen-etal-2020-generating,wang2022cross,wang2022medical} are commonly used to learn and encode priori domain information.

\noindent
{\bf Unpaired medical report generation.}
The only work that addresses the task of unpaired medical report generation is KGAE~\cite{liu2021auto}.
This work uses a pre-constructed knowledge graph, together with image labels and report labels (regarding several thoractic pathologies), to map images and reports to a shared embedding space.
Notably, the image and report datasets must follow the same labeling schema.

\noindent
{\bf Unpaired image captioning.}
In the domain of natural images, there is an abundance of auxiliary data and pre-trained models available to establish connections between vision and language.
Common approaches include object-centric methods that rely on external annotated sources~\cite{hendricks2016deep,venugopalan2017captioning} and the use of pre-trained models such as object detectors and classifiers~\cite{feng2019unsupervised,gu2019unpaired,laina2019towards,liu2021exploring2,meng2022object}.
Conversely, in the medical domain, object-based approaches are unsuitable due to the primary focus on diagnosis. 
Locating abnormalities is challenging due to their size, distribution, relation to other organs, and the limited availability of data.

\noindent
{\bf Unsupervised machine translation.}
The task of translating text between languages without relying on parallel corpora or human supervision, has also gained attention in recent years.
Advancements occurred thanks to initialization schemes and the back-translation approach, which rely on generating pseudo-language pairs~\cite{artetxe2018unsupervised,lample2018unsupervised,lample-etal-2018-phrase} or extracting them from a real corpus~\cite{wu-etal-2019-extract}.
Similar approaches have also been employed in unsupervised speech-to-speech translation~\cite{nachmani2023translatotron}.

\section{Method}
\label{sec:method}

Our goal is to create a model capable of generating medical reports for X-ray images, using two separate datasets--one for reports and the other for images.
The grand challenge lies in the absence of paired image-report data during training; in other words, there is no direct correlation between a report from one dataset and an image from the other.
We propose to tackle the problem from four perspectives, each addressing a different aspect of the challenge, which collectively provides a coherent solution.
First, we establish cross-modality relationships through a new concept: pseudo-reports.
Second, encoding images and reports using similar procedures and relying on joint global and local representations will enable mapping between the embedding spaces.
Third, cycle-consistent mapping functions will learn how to transform an image representation into a report representation.
Finally, by using only reports, a decoder is trained to generate medical reports.

Our method, illustrated in Figure~\ref{fig:method}, incorporates these four ideas.
Its preprocessing procedure generates pseudo-reports. 
Additionally, the model comprises two encoders: one dedicated to image encoding and the other to report encoding. 
Furthermore, it integrates two mapping networks: one responsible for converting image representations into report representations ($I2R$), and another for the reverse transformation ($R2I$), thereby implementing the third idea. 
Finally, the model incorporates a report decoder, encapsulating the fourth idea.

During inference, an input image is encoded by the image encoder.
The encoded image representation is then mapped to the report space by a mapping function ($I2R$).
Given this transformed representation, the decoder outputs a report that correlates to the input image.

Below, we use the following notations:
$\mathcal{D}_I$ is the image datasets, $i \in \mathcal{D}_I$ is an image and $z_i$ is $i$'s representation.
The pseudo-report of image $i$ is $\phi(i)$ and its representation is $z_{\phi(i)}$.
$\mathcal{D}_R$ is the report datasets, $r \in \mathcal{D}_R$ is a report and $z_r$ is $r$'s representation.

\subsection{Preprocessing: Pseudo-report generation}
\label{sec:pseudo}

Given an image, the basic challenge lies in establishing a relationship between its representation and a relevant report representation, despite the lack of paired data.
We propose a simple approach to solve this problem: generate a {\em pseudo-report} by employing available domain information.
While this report may not be a detailed report, it provides an image-report relationship that shares semantic similarities.
We will demonstrate our ability to utilize such pairs for guiding the mapping functions.

The pseudo-reports are pieces of text
generated by leveraging domain-specific information available for images.
For example, if labels indicating the presence or absence of specific pathologies are accessible, we incorporate this information into the pseudo-reports. 
If reports are available in languages other than English, we rely on an automatic translator, although it may not be optimized for the medical domain, to produce these pseudo-reports. 
In both cases, inaccuracies in terms of deviations from human-written reports and the level of detail may be introduced. 
 However, these pseudo-reports suffice for our goal of simply maintaining similarity in high-level semantic content.

To understand the differences between a report and a pseudo-report, a 
report from a study concerning atelectasis and cardiomegaly could be
{\em "Low lung volumes and distended bowel as described on concurrent CT abdomenpelvis. 
There are patchy opacities suggesting minor dependent bibasilar atelectasis. 
There is persistent cardiomegaly. 
There is no pneumothorax or pleural effusion." }
Our related pseudo-report could be {\em "There is cardiomegaly. 
There is atelectasis. No pleural effusion. No pneumothorax."}
Notably, the latter provides partial information and differs in style.

\noindent
\textbf{Cross-modality constraint.}
During model training, these pseudo-reports serve as a cross-modality constraint, to encourage similarity between matching global representations.
Given an image and its corresponding pseudo-report, our objective is to ensure that the transformed (i.e. mapped by $I2R)$ global representation of an image closely aligns with that of its pseudo-report, and similarly that the transformed (i.e. mapped by $R2I)$ global representation of a pseudo-report closely resembles that of the original image.
This constraint is implemented by the following loss:
\begin{equation}
\begin{aligned}
    & D_{RS} = \Delta_{cnt} \big( I2R(z_i), z_{\phi(i)}, \{z_{\phi(j)} | {\scriptstyle j\neq i}\} \big) \\
    & D_{IS} = \Delta_{cnt} \big( z_i, R2I(z_{\phi(i)}), \{R2I(z_{\phi(j)}) | {\scriptstyle j\neq i}\} \big) \\
    & L_{cm} = \frac{1}{2 | \mathcal{D}_I |} \sum _{ i \in \mathcal{D}_I, j \neq i}
    \Big( D_{RS} \, + D_{IS} \Big) \, ,
\end{aligned}
\label{eq:cm}
\end{equation}
where $\Delta_{cnt}(a,b,C)$ measures the dissimilarity between the global representations by employing a contrastive loss, considering $a$ and $b$ a positive pair and $C$ the set of negatives.

\subsection{Report \& image encoding}
\label{sec:encoding}

Our aim when processing a report or an image is to extract valuable representations that encapsulate the subtleties of the data and can later be decoded into a coherent and informative report.
In the training phase, the report representation will serve to reconstruct the input report through auto-encoding.
However, during inference, the image representations will be utilized to generate a report.

A key observation in chest X-ray images and reports is that indicators of abnormalities are often subtle, occupying only a small portion of the image or a few words in the report~\cite{dawidowicz2023limitr}. 
However, detecting these abnormalities is the core goal in this domain. 
Hence, we propose to employ two levels of representation: the local level, which focuses on image patches and report words, and the global level, encompassing the entirety of the image and report. 
While the global representation should capture the essence, such as the presence of pathologies, the local representation delves into a multitude of details that appear in the image or the report.

Typically, encoders extract local features. 
We propose to generate the global representation by a weighted sum of the local representations, utilizing self-attention~\cite{lin2017structured}. 
Hence, the influence of distinct local representations on the global representation varies based on their content and significance. 
In addition, the connection between global and local representations ensures that any  loss imposed on the global representation propagates to the local representations.

 \begin{figure*}[t]
\centering
\includegraphics[width=1.0\linewidth]{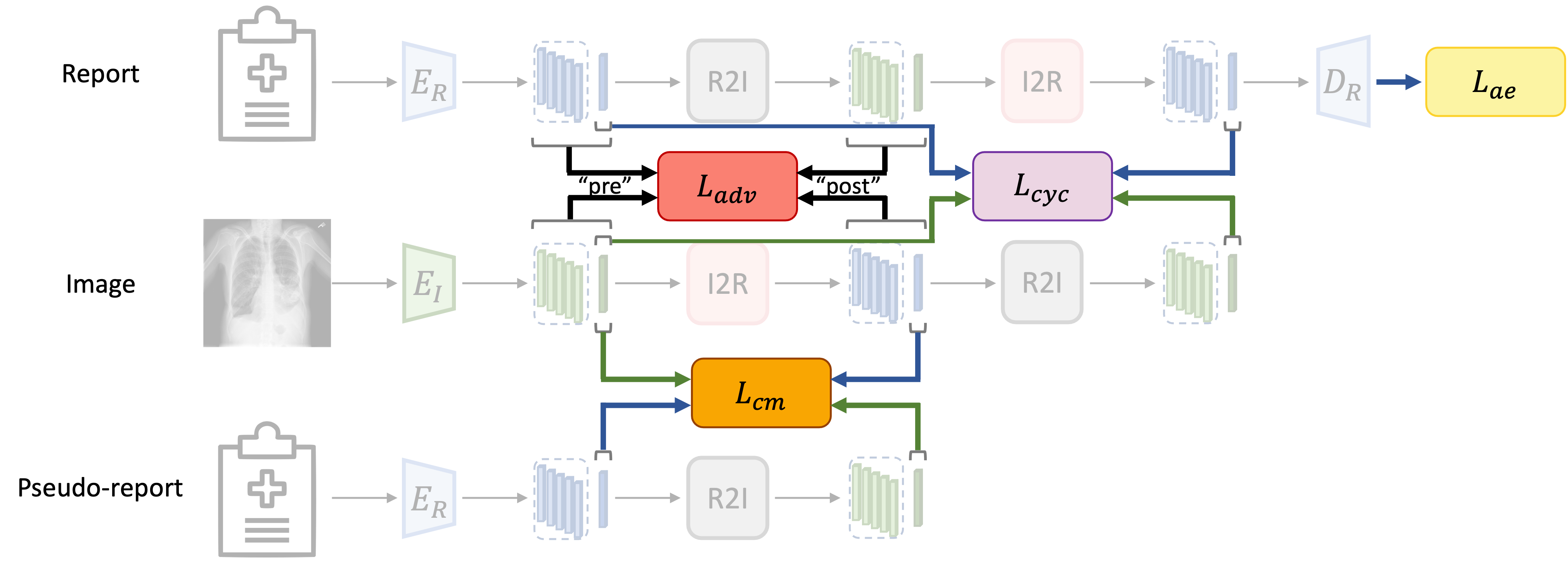}
\caption{{\bf Training objectives.} 
Our training involves four distinct objectives, each corresponding to a different loss.
The \textcolor{blockyellow}{auto-encoding loss (yellow)} focuses on accurately reconstructing the input report.
The \textcolor{blockviolet}{cycle loss (violet)} ensures cycle consistency in the $I2R$ and $R2I$ mappings.
The \textcolor{blockred}{adversarial loss (red)} ensures that the representations exhibit the same distribution before and after the mapping. 
Lastly, the \textcolor{blockorange}{cross-modal loss (orange)} aims to constrain the mapping by ensuring that pseudo-reports containing information related to an input image and the corresponding image have similar global representations.
}
\label{fig:objectives}
\end{figure*}

\subsection{Cycle-consistent mapping}
\label{sec:mapping}

The mapping, {\em I2R}, is the key component during inference, as it transforms image representations into report representations.
Since image-report pair correspondence is unavailable, we propose to train this module using a cycle-consistency constraint.
This is done as follows:
First, we employ an additional transformation that converts report representations into image representations,  $R2I$. 
Hence, we create a cycle where an image representation is mapped to a report representation by the $I2R$ module and then back by the $R2I$ module.
Second, to ensure that $I2R$ outputs valid report representations, we apply these modules in the reversed order as well, starting from a report representation, demanding cycle consistency for that modality too.
Finally, to further promote the similarity of distributions of the spaces before and after the transformations, we employ adversarial training.
We elaborate hereafter.

\noindent
\textbf{Cycle objective.}
Our approach requires consistency in two cycles: from image to image through report and vice versa.
For both cycles, we demand that the reconstructed representation $\hat{z_i}$  (/$\hat{z_r}$) resembles the original representation $z_i$ (/$z_r$).
For instance,  given an image representation $z_i$, its reconstruction $\hat{z_i}$ is attained by applying both mapping functions, $I2R$ and $R2I$, sequentially, i.e. $\hat{z_i}= R2I(I2R(z_i))$.
Hence, the loss is:
\begin{equation}
\begin{aligned}
    L_{cyc} = & \frac{1}{| \mathcal{D}_I |} \sum _{i \in \mathcal{D}_I, j \neq i} \Delta_{cnt} (z_i, \hat{z_i}, \{\hat{z_j}\}) \, + \\
    & \frac{1}{| \mathcal{D}_R |} \sum _{r \in \mathcal{D}_R, s \neq r} \Delta_{cnt} (z_r, \hat{z_r}, \{\hat{z_s}\}) \, ,
\end{aligned}
\end{equation}
where $\Delta_{cnt}(a,b,C)$ quantifies the dissimilarity between the global representations employing contrastive loss, as in Eq.~\ref{eq:cm}. 

\noindent
\textbf{Adversarial regularization.}
To ensure the intended performance of our decoder during inference, its input should  resemble the training data. 
In our case, this implies that the transformed image representations $I2R(z_i)$ should appear as though they were sampled from the report space.
Toward this end, we propose to employ adversarial training, which aims to align embedding spaces, making two spaces indistinguishable.
For that purpose, we utilize an auxiliary neural network that functions as a discriminator during training~\cite{ganin2016domain}.

Throughout training, the discriminator's objective is to distinguish between the embedding vectors from the source space (prior to mapping) and the target space (after mapping).
Given the representations $z_i$ and $z_r$, along with their respective mappings $I2R(z_i)$ and $R2I(z_r)$, the discriminator attempts to classify $z_i$ \& $z_r$ into one class (a pre-mapping class) and $I2R(z_i)$ \& $R2I(z_r)$ into another class (a post-mapping class).
The encoders and the mapping modules ($I2R$ and $R2I$) are trained to fool the discriminator, promoting indistinguishable representations. 
The discriminator produces four predicted probabilities, denoted as $p_{disc}(0|z_r)$ and $p_{disc}(0|z_i)$, representing the probability of the original representation to belong to the pre-mapping space, where $p_{disc}(1|R2I(z_r))$ and $p_{disc}(1|I2R(z_i))$ represent the probability of the transformed representations to belong to the post-mapping space.
The discriminator is trained to minimize the following loss function:
\begin{equation}
\begin{aligned}
    L_{disc} ^{(r)} = - \frac{1}{| \mathcal{D}_R |} \sum _{r \in \mathcal{D}_R} \log \big( p_{disc} (0|z_r) \big) \, + \\
    - \frac{1}{| \mathcal{D}_I |} \sum _{i \in \mathcal{D}_I} \log \big( p_{disc} (1|I2R(z_i)) \, .
\end{aligned}
\end{equation}
Here, we maximize the likelihood that the discriminator classifies the report representations as belonging to the pre-mapping space (the $0$ class) and that the transformed image representation belongs to the post-mapping space (the $1$ class). 
We similarly compute $L_{disc}^{(i)}$ for discriminating $z_i$ and $R2I(z_i)$.
Then, $L_{disc} = L_{disc}^{(r)} + L_{disc}^{(i)}$.
 
Recall that our model aims to deceive the discriminator.
Hence, it has a loss $L_{adv}$ that has the same structure as $L_{disc}$, but with the labels $0$ and $1$ swapped.
In other words, it aims to make the discriminator classify the representations incorrectly.

\subsection{Report decoding}
\label{sec:decoding}

The decoder shall translate latent representations, either of a report or an image, into a coherent report.
Due to the lack of image-report pairs, we only train it to generate text from representations of textual reports.
The generation process is learned through the auto-encoding of an input report, a task that requires only a dataset of reports.

Nevertheless, the auto-encoding task is prone to overfitting, and such models often learn to copy the input word by word.
To increase the generalization capabilities of our decoder, we apply distortions to its input during training by masking out vector representations (input dropout).
We apply dropout to the decoder's input for local representations, which express the many details appearing in every report, but not to the global ones which aim to capture high-level semantics.

\noindent
\textbf{Auto-encoding objective.}
The auto-encoding procedure aims to reconstruct input reports.
Given a report $r$, its reconstruction $\hat{r}$ should be as identical as possible to $r$.
For that purpose, we formulate the following loss function:
\begin{equation}
    L_{ae} = \frac{1}{| \mathcal{D}_R |} \sum _{r \in \mathcal{D}_R} \Delta_{ce} (r, \hat{r}) \, ,
\end{equation}
where $\Delta_{ce}(\cdot,\cdot)$ measures the dissimilarity between the two reports, calculated as the summation of token-level cross-entropy. 
In our experiments, we found that a dropout value of probability $p=0.9$ yielded good results.

\vspace{0.1in}
\noindent
\textbf{Overall loss.}
The final training objective is the sum of all the previously mentioned objectives (Figure~\ref{fig:objectives}).
For the report generation model it is:
\begin{equation}
\begin{aligned}
    L = \gamma_1 \cdot L_{cm} + \gamma_2 \cdot L_{cyc} + \gamma_3 \cdot L_{adv} + \gamma_4 \cdot L_{ae}.
\end{aligned}
\end{equation}
The parameters $\gamma_1, \dots, \gamma_4$ are hyper-parameter weights.
We set them to $\gamma_1=3, \gamma_2=1, \gamma_3=0.25, \gamma_4=1.5$.
In practice, the loss is computed for a single batch every training iteration.

\begin{table*}[t]
\centering
\setlength\tabcolsep{4pt}
\begin{tabular}{m{1.15in} |
>{\centering}m{0.3in} >{\centering}m{0.3in} >{\centering}m{0.3in} >{\centering}m{0.3in} >{\centering}m{0.3in} >{\centering}m{0.3in} | 
>{\centering}m{0.3in} >{\centering}m{0.3in} >{\centering}m{0.3in} >{\centering}m{0.32in} >{\centering}m{0.3in} >{\centering\arraybackslash}m{0.3in}}
\multirow{2}{*}{Method} & \multicolumn{6}{c|}{MIMIC-CXR} & \multicolumn{6}{c}{IU X-ray} \\[0.01in]
& B-1 & B-2 & B-3 & B-4 & M & R-L & B-1 & B-2 & B-3 & B-4 & M & R-L \\[0.01in] \hline
\vspace{0.01in}
KGAE {\small CheXpert} & 0.221 & 0.144 & 0.096 & \textbf{0.062} & 0.097 & 0.208 & 0.417 & 0.263 & 0.181 & 0.126 & 0.149 & 0.318 \\
MedCycle {\small CheXpert} & \textbf{0.309} & \textbf{0.167} & \textbf{0.098} & 0.061 & \textbf{0.115} & \textbf{0.216} & \textbf{0.461} & \textbf{0.290} & \textbf{\underline{0.201}} & \textbf{\underline{0.143}} & \textbf{0.182} & \textbf{0.332} \\[0.01in]
\hline\hline
MedCycle {\small PadChest} & 0.349 & \textbf{\underline{0.195}} & \textbf{\underline{0.115}} & \textbf{\underline{0.072}} & 0.128 & 0.239 & \textbf{\underline{0.479}} & \textbf{\underline{0.291}} & 0.198 & 0.140 & \textbf{\underline{0.197}} & \textbf{\underline{0.360}} \\[0.01in]
MedCycle \hspace{0.2in}{\small Xlate} & \textbf{\underline{0.352}} & 0.194 & 0.114 & 0.070 & \textbf{\underline{0.132}} & \textbf{\underline{0.241}} & 0.432 & 0.277 & 0.186 & 0.128 & 0.188 & 0.325 \\[0.01in]
\end{tabular}
\caption{\textbf{Quantitative evaluation, NLG metrics.}
Our MedCycle results outperform those of KGAE~\cite{liu2021auto} across all datasets \&  most of natural language generation metrics: BLEU (B), METEOR (M) \& ROUGE-L (R-L), when trained on images from CheXpert dataset.
Our results improve when the model is trained on images from PadChest, a dataset that cannot be supported by KGAE.
For the IU X-ray dataset, our model was not exposed to any
data from the dataset during training, whereas KGAE uses its reports.}
\label{table:quant_nlg}
\end{table*}

\section{Experimental Results}
\label{sec:results}

{\bf Datasets.}
We trained our model using chest X-ray images from the CheXpert dataset~\cite{irvin2019chexpert} or from the PadChest dataset~\cite{bustos2020padchest}, while the training reports were obtained from the MIMIC-CXR dataset~\cite{johnson2019mimic}. 
For performance evaluation, we utilized test sets from MIMIC-CXR and IU X-ray~\cite{demner2016preparing}. 
Our experimental configuration closely aligns with that of~\cite{liu2021auto}, with the distinction that they additionally trained on reports from the IU X-ray dataset. 
We followed the same report preprocessing steps as~\cite{chen-acl-2021-r2gencmn,liu2021auto}, which involved filtering out reports lacking a findings section. 
Importantly, no paired samples were available between the CheXpert (/PadChest) dataset and either the MIMIC-CXR or IU X-ray datasets.

The above datasets are elaborated upon in Appendix~\ref{sec:appendix_datasets}.
Generally speaking, each image is  CheXpert is assigned with multi-labels for $14$ potential diagnosis classes and the corresponding medical reports are not publically available.
PadChest is a substantial Spanish dataset of chest X-ray images, associated with medical reports. 
Each image is labeled according to $174$ possible radiographic findings.
MIMIC-CXR and IU X-ray are English dataset of chest radiographs, containing  images and corresponding reports.

\noindent
{\bf Evaluation metrics.}
We evaluate our model on two aspects: the quality of the generated language ({\em NLG}) and its clinical efficacy ({\em CE}).
For NLG evaluation, we employ the BLEU~\cite{papineni-etal-2002-bleu}, METEOR~\cite{banerjee-lavie-2005-meteor}, and ROUGE-L~\cite{lin-2004-rouge} metrics to measure the similarity between the generated reports and the ground-truth.
For CE assessment, we utilize the CheXpert~\cite{irvin2019chexpert} model to attribute $14$ diagnosis classes related to thoracic diseases and support devices. 
We then calculate precision, recall \& F1 score in comparison to the ground-truth labels.

 \begin{figure*}[t]
\centering
\begin{tabular}{>{\centering}m{0.25\linewidth} m{0.34\linewidth} m{0.34\linewidth}}
    \includegraphics[width=1\linewidth]{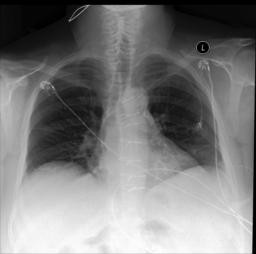} &

    
    {\small
    There are low lung volumes. 
    \ctext[RGB]{149,182,213}{The lungs are clear}. 
    There is \ctext[RGB]{192,185,191}{no pleural effusion} or \ctext[RGB]{155,208,183}{pneumothorax}. 
    The \ctext[RGB]{252,241,192}{cardiomediastinal silhouette is unremarkable}. 
    Left central line terminates in the right atrium. 
    \ctext[RGB]{255,214,165}{Median sternotomy wires} and mediastinal clips are noted. 
    A calcified lymph node is noted in the AP window.}
    &
    {\small
    Compared to prior exam from. 
    \ctext[RGB]{149,182,213}{The lungs are clear}. 
    There is \ctext[RGB]{192,185,191}{no pleural effusion} or \ctext[RGB]{155,208,183}{pneumothorax}. 
    The \ctext[RGB]{252,241,192}{cardiomediastinal silhouette is normal}.  
    \ctext[RGB]{255,214,165}{Median sternotomy wires} are intact.}
 \\
      (a) Input image & \centering (b) Ground-truth report & \centering\arraybackslash (c) Our report \\
\end{tabular}
\vspace{-0.1in}
\caption{{\bf Qualitative evaluation.} 
Our model-generated report (c) contains similar information to the ground-truth report (b).
It indicates the \ctext[RGB]{149,182,213}{lung clarity}, the \ctext[RGB]{252,241,192}{cardiomediastinal silhouette's} state, the appearance of \ctext[RGB]{255,214,165}{median sternotomy wires}, and rules out \ctext[RGB]{192,185,191}{pleural effusion} \& \ctext[RGB]{155,208,183}{pneumothorax}.
}
\label{fig:qualitative}
\end{figure*}

\noindent
{\bf Quantitative evaluation.}
Table~\ref{table:quant_nlg} provides a NLG comparative analysis between our method and KGAE~\cite{liu2021auto}, which is the only work addressing the same task. 
When trained on images from CheXpert, our model outperforms KGAE's across all metrics, except for BLEU-4 on a single dataset where it remains competitive.
When trained on images from PadChest, instead of CheXpert, our results improve on the same test datasets (MIMIC-CXR \& IU X-ray). This can be explained by the more detailed data on additional pathologies and the availability of Spanish reports.
Notably,~\cite{liu2021auto} is unable to utilize PadChest for training, due to its distinct labeling schema compared to the report dataset MIMIC-CXR ($174$ vs. $14$ labels).
We explore two approaches for generating pseudo-reports for PadChest: extracting the provided labels (PadChest) or translating the accompanying Spanish reports into English using a general translator (Google Translate; PadChest-Xlate).

The improved results demonstrate the potential of leveraging varying datasets, extending beyond those sharing similar labeling schemas.
Moreover, our method achieves results for the IU X-ray dataset in a zero-shot manner, implying no exposure to any data from this dataset during training -- neither images nor reports.
In contrast,~\cite{liu2021auto} utilizes its reports for training.

Our reports not only resemble the ground-truth but also demonstrate higher accuracy and informativeness in extracting clinical information.
These findings are depicted in Table~\ref{table:quant_ce}.
As discussed in~\cite{chen-acl-2021-r2gencmn}, these metrics cannot be employed on IU X-ray dataset, due to its labeling schema, hence Table~\ref{table:quant_ce} focuses on MIMIC-CXR.
Remarkably, training on PadChest leads to improved results for these clinical metrics as well.

\begin{table}[t]
\centering
\setlength\tabcolsep{3pt}
\begin{tabular}{m{1.15in} |
>{\centering}m{0.5in} >{\centering}m{0.5in} >{\centering\arraybackslash}m{0.5in}}

Method & Precision & Recall & F1 \\[0.01in] \hline
\vspace{0.01in}
KGAE {\small CheXpert} & 0.214 & 0.158  & 0.156 \\
MedCycle {\small CheXpert} & \textbf{0.230} & \textbf{0.171} & \textbf{0.183} \\ \hline\hline
MedCycle {\small PadChest} & \textbf{\underline{0.237}} & 0.197 & 0.183 \\
MedCycle \hspace{0.2in}{\small Xlate} & 0.218 & \textbf{\underline{0.209}} & \textbf{\underline{0.198}} \\
\end{tabular}
\caption{\textbf{Quantitative evaluation, CE metrics.}
Our MedCycle results outperform those of KGAE in terms of clinical efficacy metrics, on MIMIC-CXR, when trained on images from CheXpert or from PadChest. 
}
\label{table:quant_ce}
\end{table}

\noindent
{\bf Qualitative evaluation.}
The comparison between our generated report and the ground truth is illustrated in Figure~\ref{fig:qualitative}. 
Notably, our generated reports contain similar information to what appears in the ground truth, such as the lung clarity, the appearance of sternotomy wires, and absence of pleural effusion or pneumothorax.

\noindent
{\bf Implementation details.}
The encoder $E_R$ is a sequence of an encoding layer~\cite{bengio2000neural} and three Transformer encoder layers~\cite{vaswani2017attention}, while the encoder $E_I$ is a sequence of a ResNet-101~\cite{he2016deep} and three Transformer encoder layers. 
The decoder $D_R$ is a sequence of three Transformer decoder layers.
Both $I2R$ and $R2I$ are implemented as a simple multi-head attention layer with $8$ heads.
For $\Delta_{cnt}$ we set the temperature value to $\tau=0.1$.
We train with a batch size of $128$ on a single NVIDIA A100 GPU.

\section{Ablation Study}
\label{sec:ablation}

\begin{table}[t]
\centering
\small
\setlength\tabcolsep{3pt}
\begin{tabular}{>{\centering}m{0.25in} | >{\centering}m{0.25in} | >{\centering}m{0.25in} | >{\centering}m{0.25in} |
>{\centering}m{0.3in} >{\centering}m{0.3in} >{\centering}m{0.3in} 
 >{\centering\arraybackslash}m{0.3in}}
$L_{cm}$ & $L_{cyc}$ & $L_{adv}$ & $L_{ae}$ & B-1 & B-4 & M & F1 \\[0.01in] \hline
\vspace{0.01in}
\checkmark & \checkmark & \checkmark & \checkmark & \textbf{0.309} & \textbf{0.061} & \textbf{0.115} & \textbf{0.183} \\
 & \checkmark & \checkmark & \checkmark & 0.255 & 0.055 & 0.103 & 0.084 \\
\checkmark &  & \checkmark & \checkmark & 0.294 & 0.060 & 0.113 & 0.145 \\
\checkmark & \checkmark & & \checkmark & 0.286 & 0.055 & 0.105 & 0.151 \\
\checkmark & \checkmark & \checkmark & & 0.000 & 0.000 & 0.003 & 0.023 \\
\end{tabular}
\caption{\textbf{Ablation study, losses.}
Every loss contributes to the overall improvement in performance across all metrics, including both language and clinical aspects.
}
\label{table:ablation_loss}
\end{table}

\noindent
{\bf Losses.}
Table~\ref{table:ablation_loss} demonstrates that optimal performance is achieved when combining all our objectives. 
Applying $L_{ae}$ is crucial for training the generation of reports; otherwise, the decoder fails to learn to produce samples from the report domain.
$L_{cm}$ plays a significant role in generating reports closely associated with the input image. 
The absence of $L_{cm}$ results in relatively poor performance, particularly in the $F1$ metric, suggesting potential inaccuracies in capturing the essential information of the data -- the pathologies.
Lastly, both $L_{cyc}$ and $L_{adv}$ contribute to further enhancing the results, as they are applied on representations of actual reports and images, rather than the pseudo-reports.

\noindent
{\bf Global \& local representations.}
Table~\ref{table:ablation_gl} illustrates the impact of employing both global and local representations as inputs for the decoder. 
Across most metrics, utilizing both representations yields better results. 
Notably, the F1 score highlights that using only one representation leads to an inadequate expression of essential data elements.
Solely relying on the global representation produces favorable results in terms of NLG metrics but at the expense of a significantly poor F1 score. 
Employing only the local representation enhances the F1 score but results in a decline in NLG metrics. 

\begin{table}[t]
\centering
\small
\begin{tabular}{c|c|cccc}
\thead{decode \\ w/global} & \thead{decode \\ w/local} & B-1 & B-4 & M & F1 \\[0.01in] \hline
\vspace{0.01in}
\checkmark & \checkmark & \textbf{0.309} & 0.061 & \textbf{0.115} & \textbf{0.183} \\
 \checkmark & & 0.307 & \textbf{0.067} & 0.111 & 0.123 \\
 & \checkmark & 0.288 & 0.061 & 0.109 & 0.152 \\
\end{tabular}
\caption{\textbf{Ablation study, global \& local representations.}
Both representations contribute to the decoding process, especially in terms of the F1 metric.
}
\label{table:ablation_gl}
\end{table}

\section{Conclusions}
This paper presents a novel approach to generate X-ray reports in an unpaired manner, eliminating the need for paired images and reports during training. 
Our method integrates four key components:
(1)~Learning a mapping function between the image and report spaces through cycle-consistency.
(2)~Creating representations based on both local and global information that suit the problem and the domain.
(3)~Learning report auto-encoding.
(4)~Generating pseudo-reports utilizing domain knowledge associated with the image dataset.
 
We show the effectiveness of our method on two different datasets, surpassing the performance of existing unpaired techniques for generating chest X-ray reports.
For instance, when trained on the same image dataset as previous methods, our approach 
improves the BLEU-1 score (language metric) by $4\%$-$9\%$, depending on the dataset and the F1 score (clinical metric) by $3\%$.
When trained on different datasets, which could not be utilized by other unpaired methods due to distinct labeling schemas, the results are further improved.

\paragraph*{Limitations.}
Our model requires the image dataset to be accompanied by relevant domain-specific information, such as pathologies or reports in some language. There are datasets that lack such information and, as a result, cannot be utilized.
Furthermore, our model generates a report based on a single input image. 
However, medical examinations often reference previous findings or compare changes in severity over time, information that might be available from another image or a summary report. 
Our model is unable to utilize this contextual information despite its significance.

\paragraph*{Ethical considerations.} 
In the medical domain data privacy is a core concern.
The datasets we employed~\cite{demner2016preparing,johnson2019mimic,bustos2020padchest} were de-identified and anonymized to ensure privacy protection. 
However, deploying such models on private datasets (e.g., hospital archives) without robust privacy measures may risk the exposure of personal and sensitive information through generated reports. 
Notably, compared to paired approaches, our unpaired methodology enhances privacy by not relying on paired patient data, mitigating potential privacy breaches.

In terms of application, automatic report generation aims to enhance patient care and alleviate the burden on healthcare providers. 
Nevertheless, the automated system remains susceptible to errors, which could result in inaccurate diagnoses. 
Considering the profound consequences of erroneous diagnoses, we advocate that such automated systems should complement radiologists rather than replace them in real-world applications.

\bibliography{bib}

\newpage
\appendix
\onecolumn


\section{Datasets}
\label{sec:appendix_datasets}

{\em CheXpert}~\cite{irvin2019chexpert} is a dataset of chest X-ray images, containing $224,316$ radiographs from $65,240$ patients, collected at Stanford Hospital.
Each image is assigned with multi-labels for 14 potential diagnosis classes.
The corresponding medical reports are not publically available.

\vspace{0.05in}

{\em PadChest}~\cite{bustos2020padchest} is a substantial Spanish dataset of chest X-ray images, comprising $160,868$ radiographs from $69,882$ patients, along with their associated medical reports. 
The data was collected at San Juan Hospital.
Each image is labeled according to $174$ possible radiographic findings, $19$ diagnoses, and $104$ anatomic locations. 
Each image is associated with a report in Spanish.

\vspace{0.05in}

{\em MIMIC-CXR}~\cite{johnson2019mimic} is a large English dataset of chest radiographs, containing $377,110$ images, corresponding to $227,835$ reports performed at the Beth Israel Deaconess Medical Center.
The dataset is split into $368,960$ images ($222,758$ reports) for training, $2,991$ images ($1,808$ reports) for validation, and $5,159$ images ($3,269$ reports) for testing.

\vspace{0.05in}

{\em IU X-ray}~\cite{demner2016preparing} comprises $7,470$ chest X-ray images, each associated with one of $3,955$ reports. 
We employ the same train-validation-test split of $70\%$-$10\%$-$20\%$ as defined by~\cite{li2018hybrid}.

\vspace{0.05in}
Access to the datasets is granted directly by the dataset owners upon registration and approval, owing to their sensitivity.

\end{document}